\documentclass[10pt,twocolumn,letterpaper]{article}
\usepackage{authblk}

\usepackage[pagenumbers]{cvpr}      

\usepackage{graphicx}
\usepackage{amsmath}
\usepackage{amssymb}
\usepackage{booktabs}
\usepackage{array}
\usepackage{balance}

\usepackage[pagebackref,breaklinks,colorlinks]{hyperref}

\usepackage[capitalize]{cleveref}
\crefname{section}{Sec.}{Secs.}
\Crefname{section}{Section}{Sections}
\Crefname{table}{Table}{Tables}
\crefname{table}{Tab.}{Tabs.}

\begin{document}

\title{Meerkat Behaviour Recognition Dataset}

\author[1]{Mitchell Rogers}
\author[1]{Gaël Gendron}
\author[1]{David Arturo Soriano Valdez}
\author[1]{Mihailo Azhar}
\author[1]{Yang Chen}
\author[1]{Shahrokh Heidari}
\author[1]{Caleb Perelini}
\author[1]{Padriac O’Leary}
\author[1]{Kobe Knowles}
\author[1]{Izak Tait}
\author[2]{Simon Eyre}
\author[1]{Michael Witbrock}
\author[1]{Patrice Delmas}

\affil[1]{NAOInstitute, The University of Auckland, Auckland, New Zealand}
\affil[2]{Wellington Zoo, Wellington, New Zealand}

\maketitle

\begin{abstract}
Recording animal behaviour is an important step in evaluating the well-being of animals and further understanding the natural world. Current methods for documenting animal behaviour within a zoo setting, such as scan sampling, require excessive human effort, are unfit for around-the-clock monitoring, and may produce human-biased results. Several animal datasets already exist that focus predominantly on wildlife interactions, with some extending to action or behaviour recognition. However, there is limited data in a zoo setting or data focusing on the group behaviours of social animals. We introduce a large meerkat (Suricata Suricatta) behaviour recognition video dataset with diverse annotated behaviours, including group social interactions, tracking of individuals within the camera view, skewed class distribution, and varying illumination conditions. This dataset includes videos from two positions within the meerkat enclosure at the Wellington Zoo (Wellington, New Zealand), with 848,400 annotated frames across 20 videos and 15 unannotated videos.

\end{abstract}

\section{Introduction}
\label{sec:intro}
Computer vision offers a noninvasive alternative to currently applied animal monitoring approaches, such as GPS tracking collars \cite{HUFFMEYER2022157,Perez19}. Computer vision also solves many manual processes still used in zoological studies, such as scan sampling, to estimate how animals allocate their time to certain behavioural states throughout the day \cite{Rose_2021}. As object detection methods continue to improve and researchers progress our understanding of natural intelligence in animals, a gap has been identified in the literature for recognising intelligent behaviours from videos of animals. Therefore, we propose a novel dataset of video footage covering a comprehensive range of meerkat behaviours, as identified by experts in a zoo environment.


Meerkats (\textit{Suricata Suricatta}) are a highly social mongoose species from South Western Africa. Meerkat groups vary from two to fifty individuals with a rigid social structure where a dominant pair monopolises breeding, and all other subordinate adults help take care of the pups \cite{Scott_2014}. Meerkats can exhibit many social behaviours, such as babysitting, play fighting, social digging, mutual grooming (allogrooming), and huddling \cite{Habicher_2009}. These social behaviours extend to more complex behaviours, such as conflicts which lead to eviction from their social group \cite{Stephens_2005} and reactions to potential predators based on instincts \cite{Heuls_2022}, even when there are no risks of resource scarcity and predators.

Behavioural studies are essential for evidence-based management of zoo animals \cite{Rose_2021}. The zoo environment allows researchers to observe animal behaviour in a controlled setting. Previous animal datasets focused on the classification and detection of a wide range of animals from individual frames \cite{iNaturalist, Snapshot_serengeti, Wellington_camera, Gagne_2021}, detection of diverse behaviours across species \cite{Animal_kingdom}, and pose estimation \cite{OpenMonkeyStudio, OpenMonkeyChallenge}. Many previous datasets used wildlife camera traps \cite{Snapshot_serengeti, Gagne_2021, Wellington_camera, Wildlife_trap_depth} allowing researchers to study populations in their natural environment. However, few datasets provide behaviour identifications in a zoo environment.

Zoo environments have many different characteristics from these previous datasets, with the main difference being a fixed number of individuals, more frequent behaviours, and the ability to view the entire enclosure \cite{OpenMonkeyStudio}.


This study provides an annotated video dataset for meerkat behaviour identification in a zoo environment. This dataset tracks individuals that persist in the camera view, and classifies their actions. Using two cameras within an enclosure at Wellington Zoo (Wellington, New Zealand), we labelled 848,400 video frames of a range of meerkat behaviour based on a comprehensive behavioural ethogram across 20 videos for a total of four hours of footage, and also provided another 15 videos (three hours of footage) to support active learning approaches.

\section{Methods}
\label{sec:methods}

\begin{figure*}
  \centering
  \begin{subfigure}{0.49\linewidth}
    \includegraphics[width=\textwidth]{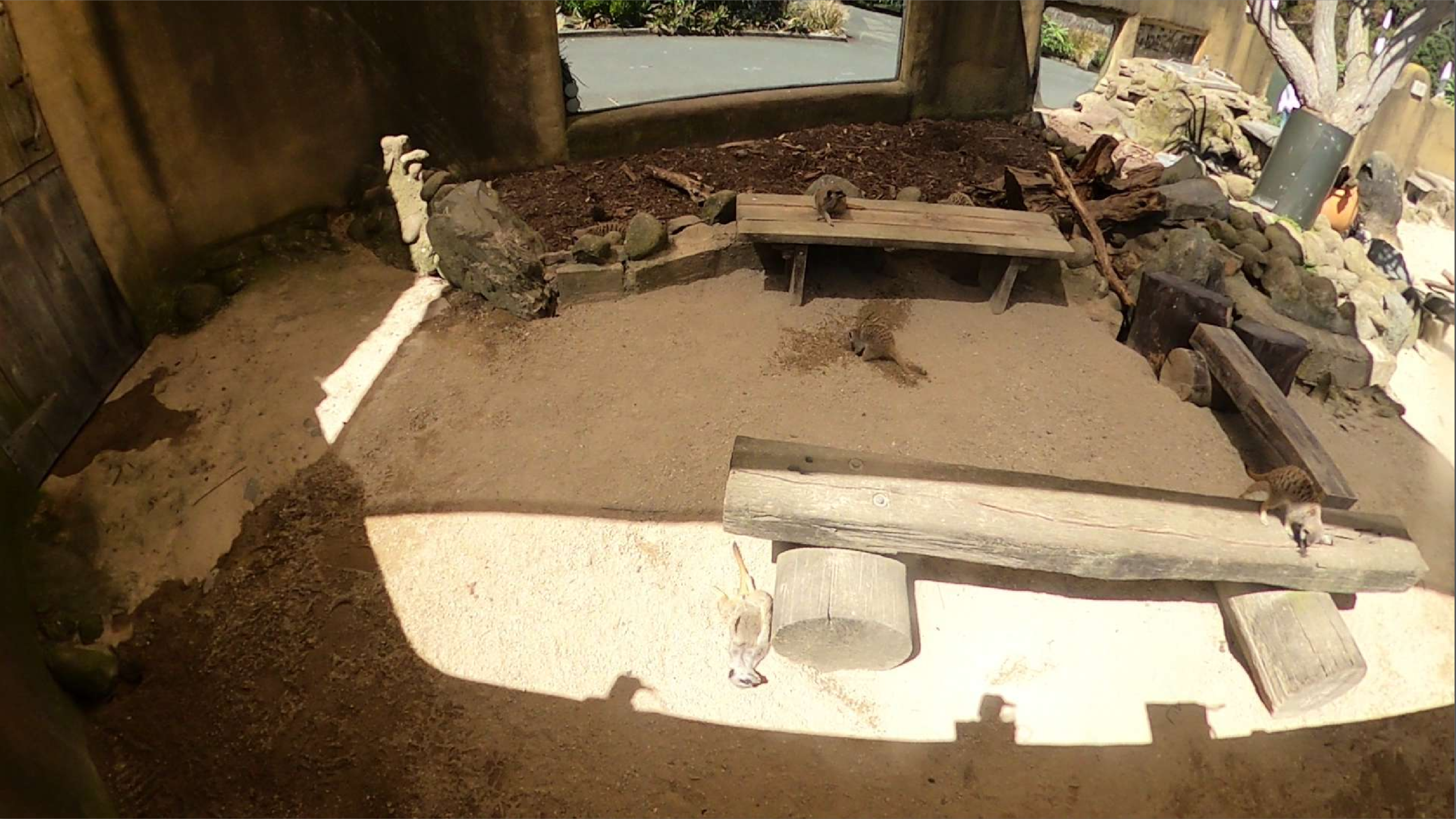}
    \caption{Camera view of the entrance and foraging area.}
    \label{fig:entrance}
  \end{subfigure}
  \hfill
  \begin{subfigure}{0.49\linewidth}
    \includegraphics[width=\textwidth]{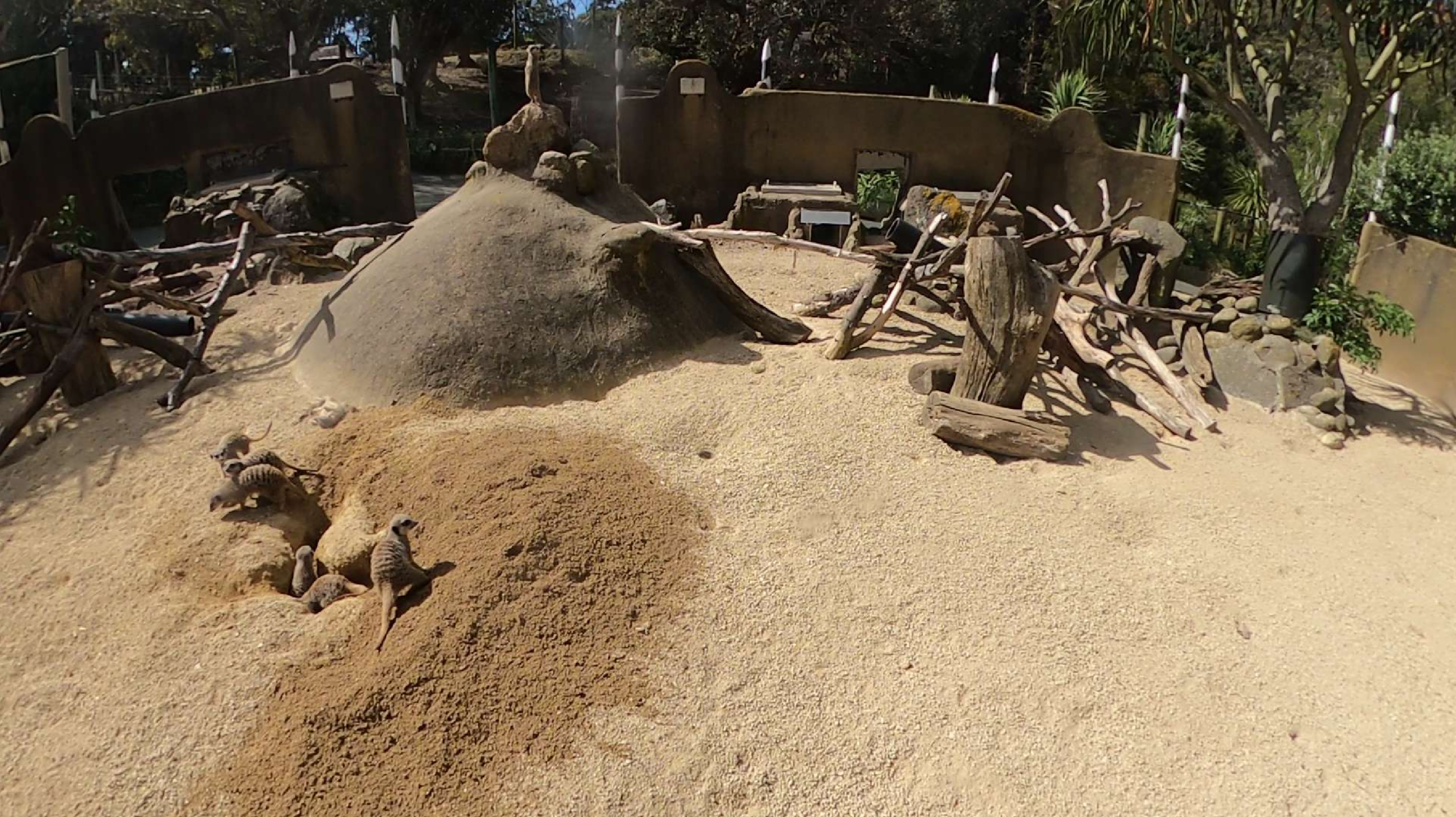}
    \caption{Camera view of the mound and backside of the enclosure.}
    \label{fig:mound}
  \end{subfigure}
  \caption{Example images of the camera views.}
  \label{fig:cams}
\end{figure*}

\subsection{Data collection}
Two cameras (GoPro® Max) were positioned on the back wall of the enclosure, one facing the replica termite mound in the centre of the enclosure and the other overlooking the foraging area and entrance to the enclosure (Figure \ref{fig:cams}). These areas were identified as those where meerkats spent significant time throughout the day. The mound (Figure \ref{fig:mound}) is a popular area for guarding (sentinel) behaviour, and at certain times of the day, meerkats congregate around the mound to bask in the sun. The foraging area (Figure \ref{fig:entrance}) is frequented around mealtimes, where meerkats dig around the loose bark for bugs or interact with objects that contain food. The meerkats will also gather around the entrance area when they expect zookepers.

Footage was collected between October and November 2022, on days with clear weather conditions. Specifically, footage was chosen from the 20th and 21st of October and the 7th of November. Each camera gathered videos at twelve-minute intervals and was screened to remove clips with guests in the enclosure. Videos with many individuals, social interactions, and other interesting behaviours were selected for the annotation. A small team of annotators then labelled the selected clips using the Computer Vision Annotation Tool (CVAT version 2.3) \cite{CVAT}. Finally, a mask was applied to blur the windows from the video frames to protect the privacy of the zoo patrons while retaining the visual presence of humans.


This dataset labelled two classes of meerkats: adult meerkats and meerkat pups. Within the data collection time frame, a litter of four meerkat pups was born, and although they were often occluded in the video frames, the annotators used a small bounding box to specify their position relative to the adults in the scene. Adults often change their behaviour in the presence of pups, often forming huddles around them to groom them. To validate the dataset and minimise biases between annotators, a different annotator for each video checked for consistent annotations (occlusion and action labels) and missed annotations. Substantial differences were observed between annotators. Some annotators had different understandings of what each behaviour looked like, leading to consistent errors across the videos. To deal with errors remaining after a second check, a third check was applied with a script to check whether the behaviours occurred in a valid location and to list the frame numbers for the behaviours that may be erroneous for a manual check.

Meerkats are difficult to identify. Previous studies identified meerkats based on the shape of the dark regions around their eyes \cite{Heuls_2022}, applying dye to their fur \cite{Thornton_2012, Scott_2014}, or invasively feeding each individual a different colour of edible glitter \cite{Scott_2017}. We or the zookeepers could not recognise individual meerkats accurately from the video footage; therefore, we were unable to provide ground-truth identification. However, each meerkat was tracked as long as they persisted within the camera view, and we know there were nine adult males and six adult females, and for the footage collected on the 7th of November, there were four young pups.

\subsection{Behavioural ethogram}
An important behaviour exhibited by meerkats is vigilance, where individuals sit in a bipedal posture and scan the horizon for predators, while others lower their guard to undertake other activities \cite{Heuls_2022}. Individuals take turns watching over others according to some irregular rota \cite{Clutton_1999} and frequently vocalise to communicate risks to the rest of the group \cite{Scott_2014}. Vigilant behaviour can be broken down into different types. Meerkats can sit with all four legs on the ground scanning for threats (vigilance pose), sit up on their hind legs with front legs against their body (guarding pose), or stand on their tiptoes supported by their tail quickly scanning for threats (raised guarding pose) \cite{Heuls_2022}. When a single meerkat stands in a vigilant stance at an elevated point, this is sentinel behaviour \cite{Heuls_2022, Scott_2014}. Specific individuals are more likely to contribute to guarding based on their nutritional status \cite{Clutton_1999}, sex, or social rank \cite{Scott_2014}.

\begin{figure*}
  \centering
  \begin{subfigure}{0.24\linewidth}
    \includegraphics[width=0.48\textwidth, height=0.5\textwidth]{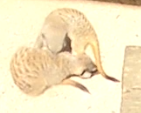}
    \includegraphics[width=0.48\textwidth, height=0.5\textwidth]{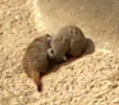}
    \caption{Allogrooming.}
    \label{fig:behaviour_allogrooming}
  \end{subfigure}
  \hfill
  \begin{subfigure}{0.24\linewidth}
    \includegraphics[width=0.48\textwidth, height=0.5\textwidth]{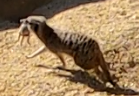}
    \includegraphics[width=0.48\textwidth, height=0.5\textwidth]{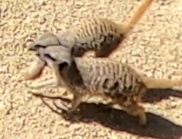}
    \caption{Carrying a pup.}
    \label{fig:behaviour_carrying}
  \end{subfigure}
  \hfill
  \begin{subfigure}{0.24\linewidth}
    \includegraphics[width=0.48\textwidth, height=0.5\textwidth]{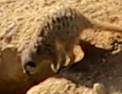}
    \includegraphics[width=0.48\textwidth, height=0.5\textwidth]{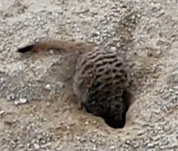}
    \caption{Digging.}
    \label{fig:behaviour_digging_burrow}
  \end{subfigure}
  \hfill
  \begin{subfigure}{0.24\linewidth}
    \includegraphics[width=0.48\textwidth, height=0.5\textwidth]{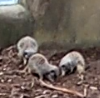}
    \includegraphics[width=0.48\textwidth, height=0.5\textwidth]{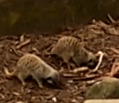}
    \caption{Foraging.}
    \label{fig:behaviour_foraging}
  \end{subfigure}
  \hfill
  \begin{subfigure}{0.24\linewidth}
    \includegraphics[width=0.48\textwidth, height=0.5\textwidth]{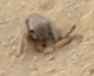}
    \includegraphics[width=0.48\textwidth, height=0.5\textwidth]{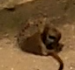}
    \caption{Grooming.}
    \label{fig:behaviour_grooming}
  \end{subfigure}
  \hfill
  \begin{subfigure}{0.24\linewidth}
    \includegraphics[width=0.48\textwidth, height=0.5\textwidth]{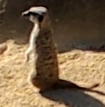}
    \includegraphics[width=0.48\textwidth, height=0.5\textwidth]{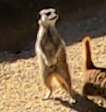}
    \caption{High sitting/standing.}
    \label{fig:behaviour_h_sitting}
  \end{subfigure}
  \hfill
  \begin{subfigure}{0.24\linewidth}
    \includegraphics[width=0.48\textwidth, height=0.5\textwidth]{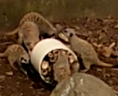}
    \includegraphics[width=0.48\textwidth, height=0.5\textwidth]{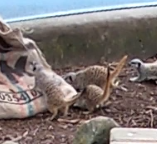}
    \caption{Interacting with a foreign object.}
    \label{fig:behaviour_object_interaction}
  \end{subfigure}
  \hfill
  \begin{subfigure}{0.24\linewidth}
    \includegraphics[width=0.48\textwidth, height=0.5\textwidth]{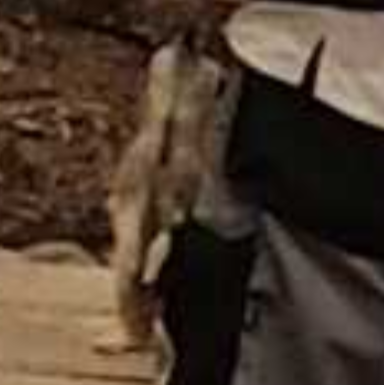}
    \includegraphics[width=0.48\textwidth, height=0.5\textwidth]{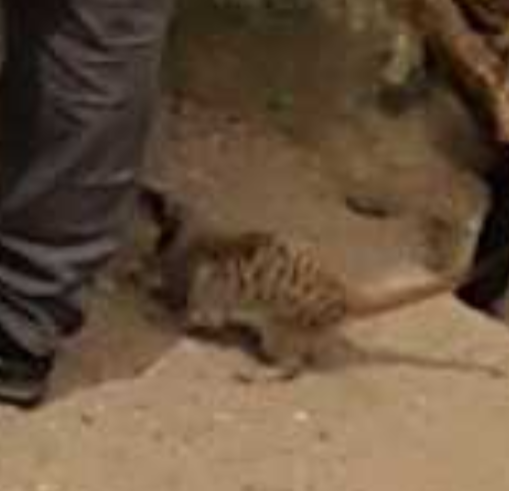}
    \caption{Interacting with a human.}
    \label{fig:human_interaction}
  \end{subfigure}
  \hfill
    \begin{subfigure}{0.24\linewidth}
    \includegraphics[width=0.48\textwidth, height=0.5\textwidth]{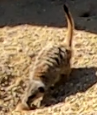}
    \includegraphics[width=0.48\textwidth, height=0.5\textwidth]{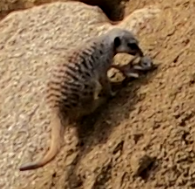}
    \caption{Interacting with a pup.}
    \label{fig:behaviour_pup_interaction}
  \end{subfigure}
  \hfill
  \begin{subfigure}{0.24\linewidth}
    \includegraphics[width=0.48\textwidth, height=0.5\textwidth]{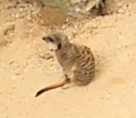}
    \includegraphics[width=0.48\textwidth, height=0.5\textwidth]{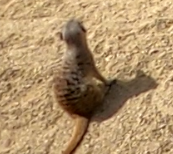}
    \caption{Low sitting/standing.}
    \label{fig:behaviour_l_sitting}
  \end{subfigure}
  \hfill
  \begin{subfigure}{0.24\linewidth}
    \includegraphics[width=0.48\textwidth, height=0.5\textwidth]{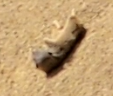}
    \includegraphics[width=0.48\textwidth, height=0.5\textwidth]{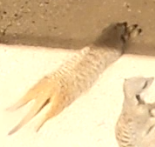}
    \caption{Lying.}
    \label{fig:behaviour_lying}
  \end{subfigure}
  \hfill
  \begin{subfigure}{0.24\linewidth}
    \includegraphics[width=0.48\textwidth, height=0.5\textwidth]{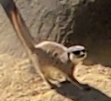}
    \includegraphics[width=0.48\textwidth, height=0.5\textwidth]{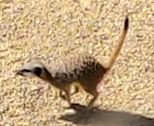}
    \caption{Moving.}
    \label{fig:behaviour_moving}
  \end{subfigure}
  \hfill
  \begin{subfigure}{0.24\linewidth}
    \includegraphics[width=0.48\textwidth, height=0.5\textwidth]{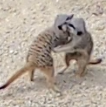}
    \includegraphics[width=0.48\textwidth, height=0.5\textwidth]{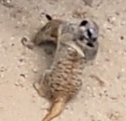}
    \caption{Playfighting.}
    \label{fig:behaviour_playfighting}
  \end{subfigure}
  \begin{subfigure}{0.24\linewidth}
    \includegraphics[width=0.48\textwidth, height=0.5\textwidth]{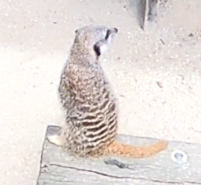}
    \includegraphics[width=0.48\textwidth, height=0.5\textwidth]{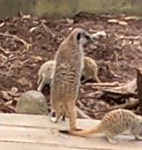}
    \caption{Raised guarding.}
    \label{fig:behaviour_raised_guarding}
  \end{subfigure}
  \begin{subfigure}{0.24\linewidth}
    \includegraphics[width=0.48\textwidth, height=0.5\textwidth]{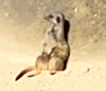}
    \includegraphics[width=0.48\textwidth, height=0.5\textwidth]{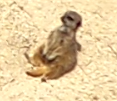}
    \caption{Sunbathing.}
    \label{fig:behaviour_sunbathing}
  \end{subfigure}
  \hfill
  \caption{Examples of the meerkat behaviours.}
  \label{fig:behaviours}
\end{figure*}

The behaviours of interest are described by the ethogram presented in Table \ref{tab:Ethogram} and example images of each behaviour are shown in Figure \ref{fig:behaviours}. An ethogram lists and defines all behaviours of interest, allowing for a reproducible way of categorising observed behaviours \cite{Rose_2021}. Each behaviour is mutually exclusive, meaning that a single individual can only have one behaviour at a time. This ethogram was adapted from Katy Scott's ethogram \cite{Scott_2014} with some changes based on the ethogram presented in Alexandra Habicher's thesis \cite{Habicher_2009}. An estimated time budget detailing the proportion of time an average meerkat would exhibit each behaviour unoccluded within the camera view was created by summing the number of frames each meerkat exhibited each respective behaviour.

\begin{table*}[t]
    \centering
    \begin{tabular}{m{3.0cm}m{2.8cm}m{10.5cm}}
        \toprule
        Behaviour & No. frames (\%) & Description\\
        \midrule
        Foraging                            & 1,202,101 (28.72\%)      & Traverses the enclosure with a ducked body and lowered tail, searching for food, digging in pursuit of food items, or biting/swallowing a food item. \\
        Raised guarding                     & 588,305 (14.06\%)      & Standing, or sunbathing in an elevated position within the enclosure.  \\
        Moving                              & 567,835 (13.57\%)     & Walks, runs, or climbs.  \\
        Low sitting/standing                & 469,214 (11.21\%)      & Sits or stands on the ground with their upper extremities on the ground.  \\
        Digging a burrow                    & 423,183 (10.22\%)     & Digs with hind legs widely spread, expelling soil between them to create or modify a burrow.  \\
        High sitting/standing               & 318,665 (7.61\%)      & Sits upright with their lower extremities or footpads touching the ground while the forelimbs are bent in front of the body. \\
        Lying/resting                       & 142,980 (3.42\%)      & Lies on the ground with their front or back side touching the ground.\\
        Interacting with a pup              & 138,730 (3.32\%)      & Within close proximity of a pup, with its attention directed towards the pup.  \\
        \shortstack[l]{Interacting with a\\foreign object}   & 	113,504 (2.71\%)      & Interacts with a stimulus object, toy or something else that does not belong within the enclosure.  \\
        Sunbathing                          & 73,748 (1.76\%)      & Vigilant lying or sitting on their backs with their bellies exposed to sunlight. \\
        Playfighting                        & 49,847 (1.19\%)       & Two or more meerkats bite, grab, or pursue one another.  \\
        Grooming                            & 37,324 (0.89\%)       & Cleans its own fur with teeth and tongue.  \\
        Carrying a pup                        & 35,417 (0.85\%)       & Moves while carrying a pup with their mouth.  \\
        Allogroom                           & 20,822 (0.50\%)       & One meerkat cleans the fur of another, either independently or reciprocally.  \\
        \shortstack[l]{Interacting with\\humans}     & 3,191 (0.08\%)       & Within close proximity of a human, with its attention directed towards this human.  \\
        \bottomrule
    \end{tabular}
    \caption{Meerkat behaviour ethogram. Based on ethograms utilised in previous studies and the behaviour observed from our camera's viewpoints.}
    \label{tab:Ethogram}
\end{table*}

\section{Discussion}
Across the 848,400 annotated frames, there were over 4.1 million unoccluded meerkat annotations, with an average of 4.9 meerkats per frame. This is far from the total number of meerkats in the enclosure at the start of the data collection period, with a litter of four pups born during the data-gathering period. Therefore, the estimated time budget does not comprehensively capture the actions of the whole population in the enclosure across the entire day, because only a portion of meerkats were within view, and the video sequences were selected to maximise activity and favourable weather conditions. In addition, the most common actions of the meerkats were foraging, taking up close to a third of their time, and vigilant behaviour (split between high sitting/standing and raised guarding), accounting for approximately a fifth of their time. The least common actions were social behaviours, playfighting, carrying pups, and allogrooming, accounting for less than 1.5\% of their time each, along with self-grooming.

The time budget of this dataset is comparable to the time budget presented in \cite{Scott_2014} of meerkats in a zoo environment visible and outside. Our dataset included a lower proportion of observations for foraging compared to other studies on captive meerkats and wild meerkats presented in \cite{Habicher_2009}. Our video selection process may have overrepresented activities, such as digging and interacting with toys containing food. Resting behaviours may be underrepresented because these behaviours may be more common outside of the camera view, such as within the den areas.

Due to the birth of the four pups before the third data collection day, we were able to observe interactions with the young pups. This also led to a change in the behaviour distribution, as before the birth of the pups, the pregnant matriarch could be frequently seen sunbathing. In one sequence, an adult meerkat carried a pup to the centre of the video and the surrounding meerkats stopped their current actions to interact with the pup, forming a huddle. This is a clear example of the causal behaviour that we aimed to capture within this dataset. Other examples of observed group behaviours were meerkats working together to dig holes and meerkats congregating close to the enclosure door when they anticipated a zookeeper entering.

\section{Conclusion}
Overall, our dataset offers valuable information about frequently observed and expected social behaviours among captive meerkats as well as less frequently observed interactions with young pups. Our future work on this annotated dataset will include automated social behaviour detection for meerkats and other species using forward/inverse reinforcement learning techniques \cite{ng2000algorithms} and the extraction of behavioural models for animals with causal structure discovery \cite{ziebart2010modeling,zhucausal,entner2010causal}, in conjunction with philosophical theories of mind \cite{jara2019theory}. One of the goals of this dataset is to push research on the modelling and explanation of complex behaviours from observations \cite{scholkopf2021toward}. Using this dataset, we aim to automatically build causal models that can represent complex interactions between animals and explain the observed behaviours.

\balance
{\small
\bibliographystyle{ieee_fullname}
\bibliography{egbib}

\begin{thebibliography}{10}\itemsep=-1pt

\bibitem{Wellington_camera}
Victor Anton, Stephen Hartley, Andre Geldenhuis, and Heiko~U Wittmer.
\newblock {Monitoring the mammalian fauna of urban areas using remote cameras
  and citizen science}.
\newblock {\em Journal of Urban Ecology}, 4(1), 03 2018.
\newblock juy002.

\bibitem{OpenMonkeyStudio}
Praneet~C. Bala, Benjamin~R. Eisenreich, Seng Bum~Michael Yoo, Benjamin~Y.
  Hayden, Hyun~Soo Park, and Jan Zimmermann.
\newblock Openmonkeystudio: Automated markerless pose estimation in freely
  moving macaques.
\newblock {\em bioRxiv}, 2020.

\bibitem{Clutton_1999}
T.~H. Clutton-Brock, M.~J. O'Riain, P.~N.~M. Brotherton, D. Gaynor, R. Kansky,
  A.~S. Griffin, and M. Manser.
\newblock Selfish sentinels in cooperative mammals.
\newblock {\em Science}, 284(5420):1640--1644, 1999.

\bibitem{CVAT}
CVAT.ai Corporation.
\newblock Computer vision annotation tool (cvat), 3 2023.

\bibitem{entner2010causal}
Doris Entner and Patrik~O Hoyer.
\newblock On causal discovery from time series data using fci.
\newblock {\em Probabilistic graphical models}, pages 121--128, 2010.

\bibitem{Gagne_2021}
Crystal Gagne, Jyoti Kini, Daniel Smith, and Mubarak Shah.
\newblock Florida wildlife camera trap dataset.
\newblock {\em CoRR}, abs/2106.12628, 2021.

\bibitem{Habicher_2009}
Alexandra Habicher.
\newblock {\em Behavioural Cost Minimisation and Minimal Invasive
  Blood-Sampling in Meerkats (S. suricatta, Herpestidae)}.
\newblock PhD thesis, Universit{\"a}t zu K{\"o}ln, 2009.

\bibitem{Wildlife_trap_depth}
Timm Haucke and Volker Steinhage.
\newblock Exploiting depth information for wildlife monitoring, 2021.

\bibitem{Heuls_2022}
Florian~D. Huels and Angela~S. Stoeger.
\newblock Sentinel behavior in captive meerkats (suricata suricatta).
\newblock {\em Zoo Biology}, 41(1):10--19, 2022.

\bibitem{HUFFMEYER2022157}
Audra~A. Huffmeyer, Jeff~A. Sikich, T.~Winston Vickers, Seth~P.D. Riley, and
  Robert~K. Wayne.
\newblock First reproductive signs of inbreeding depression in southern
  california male mountain lions (puma concolor).
\newblock {\em Theriogenology}, 177:157--164, 2022.

\bibitem{jara2019theory}
Julian Jara-Ettinger.
\newblock Theory of mind as inverse reinforcement learning.
\newblock {\em Current Opinion in Behavioral Sciences}, 29:105--110, 2019.

\bibitem{ng2000algorithms}
Andrew~Y Ng, Stuart Russell, et~al.
\newblock Algorithms for inverse reinforcement learning.
\newblock In {\em Icml}, volume~1, page~2, 2000.

\bibitem{Animal_kingdom}
Xun~Long Ng, Kian~Eng Ong, Qichen Zheng, Yun Ni, Si~Yong Yeo, and Jun Liu.
\newblock Animal kingdom: A large and diverse dataset for animal behavior
  understanding.
\newblock In {\em Proceedings of the IEEE/CVF Conference on Computer Vision and
  Pattern Recognition (CVPR)}, pages 19023--19034, June 2022.

\bibitem{Perez19}
Rodrigo Nu{\~{n}}ez-Perez and Brian Miller.
\newblock {\em Movements and Home Range of Jaguars (Panthera onca) and Mountain
  Lions (Puma concolor) in a Tropical Dry Forest of Western Mexico}, pages
  243--262.
\newblock Springer International Publishing, 2019.

\bibitem{Rose_2021}
Paul~E. Rose and Lisa~M. Riley.
\newblock Conducting behavioural research in the zoo: A guide to ten important
  methods, concepts and theories.
\newblock {\em Journal of Zoological and Botanical Gardens}, 2(3):421--444,
  2021.

\bibitem{scholkopf2021toward}
Bernhard Sch{\"o}lkopf, Francesco Locatello, Stefan Bauer, Nan~Rosemary Ke, Nal
  Kalchbrenner, Anirudh Goyal, and Yoshua Bengio.
\newblock Toward causal representation learning.
\newblock {\em Proceedings of the IEEE}, 109(5):612--634, 2021.

\bibitem{Scott_2014}
Katy Scott.
\newblock {\em Behaviour and Endocrinology of Meerkats in Zoos}.
\newblock PhD thesis, University of Exeter, 2014.

\bibitem{Scott_2017}
Katy Scott, Michael Heistermann, Michael~A. Cant, and Emma I.~K. Vitikainen.
\newblock Group size and visitor numbers predict faecal glucocorticoid
  concentrations in zoo meerkats.
\newblock {\em Royal Society Open Science}, 4(4):161017, 2017.

\bibitem{Stephens_2005}
P. A. Stephens, A. F. Russell, A. J. Young, W. J. Sutherland, and T. H.
  Clutton‐Brock.
\newblock Dispersal, eviction, and conflict in meerkats (suricata suricatta):
  An evolutionarily stable strategy model.
\newblock {\em The American Naturalist}, 165(1):120--135, 2005.
\newblock PMID: 15729644.

\bibitem{Snapshot_serengeti}
Alexandra Swanson, Margaret Kosmala, Chris Lintott, Robert Simpson, Arfon
  Smith, and Craig Packer.
\newblock Snapshot serengeti, high-frequency annotated camera trap images of 40
  mammalian species in an african savanna.
\newblock {\em Scientific Data}, 2(1):150026, Jun 2015.

\bibitem{Thornton_2012}
Alex Thornton and Jamie Samson.
\newblock Innovative problem solving in wild meerkats.
\newblock {\em Animal Behaviour}, 83(6):1459--1468, 2012.

\bibitem{iNaturalist}
Grant Van~Horn, Oisin Mac~Aodha, Yang Song, Yin Cui, Chen Sun, Alex Shepard,
  Hartwig Adam, Pietro Perona, and Serge Belongie.
\newblock The inaturalist species classification and detection dataset, 2017.

\bibitem{OpenMonkeyChallenge}
Yuan Yao, Abhiraj Mohan, Eliza Bliss-Moreau, Kristine Coleman, Sienna~M.
  Freeman, Christopher~J. Machado, Jessica Raper, Jan Zimmermann, Benjamin~Y.
  Hayden, and Hyun~Soo Park.
\newblock Openmonkeychallenge: Dataset and benchmark challenges for pose
  tracking of non-human primates.
\newblock {\em bioRxiv}, 2021.

\bibitem{zhucausal}
Shengyu Zhu, Ignavier Ng, and Zhitang Chen.
\newblock Causal discovery with reinforcement learning.
\newblock In {\em International Conference on Learning Representations}, 2019.

\bibitem{ziebart2010modeling}
Brian~D Ziebart, J~Andrew Bagnell, and Anind~K Dey.
\newblock Modeling interaction via the principle of maximum causal entropy.
\newblock {\em Proceedings of the 27th International Conference on
  International Conference on Machine Learning}, page 1255–1262, 2010.

\end{thebibliography}
}

\end{document}